%% file: paper.tex
\documentclass{article} 
\usepackage{iclr2025_conference,times}
\iclrfinalcopy 

\input{math_commands.tex}

\usepackage{hyperref}
\usepackage{url}

\usepackage{times}
\usepackage{soul}
\usepackage[utf8]{inputenc}
\usepackage[small]{caption}
\usepackage{graphicx}
\usepackage{amsmath}
\usepackage{amsthm}
\usepackage{booktabs}
\usepackage{algorithm}
\usepackage{algorithmic}
\usepackage[switch]{lineno}
\usepackage{array}
\usepackage{tabularx}

\usepackage{times}
\usepackage{epsfig}
\usepackage{graphicx}
\usepackage{amsmath}
\usepackage{amssymb}
\usepackage{booktabs}
\usepackage{float}
\usepackage{xcolor}
\usepackage{multirow}

\usepackage{comment}

\usepackage[utf8]{inputenc}
\usepackage{pgfplots}
\DeclareUnicodeCharacter{2212}{−}
\usepgfplotslibrary{groupplots,dateplot}
\usetikzlibrary{patterns,shapes.arrows}
\pgfplotsset{compat=newest}

\title{LoRA Diffusion: Zero-shot LoRA Synthesis for Diffusion Model Personalization}

\author{
Ethan Smith\thanks{Equal contribution.} \\
Leonardo AI Research Lab \\
North Sydney, NSW, Australia \\
\texttt{ethan@leonardo.ai} \\
\And
Rami M. Seid\footnotemark[1] \\
Lucid Simulations, Corp. \\
\texttt{rami@lucidsim.co} \\
\And
Alberto Hojel\footnotemark[1] \\
Lucid Simulations, Corp. \\
\texttt{alberto@lucidsim.co} \\
\And
Paramita Mishra  \\
Precigenetics Inc \\
United States \\
\texttt{parm@precigenetics.com}
\And
Jianbo Wu \\
SimpleBerry Research Lab \\
University of California, Merced \\
\texttt{jwu323@ucmerced.edu}
}

\begin{document}

\maketitle

\begin{figure}[htbp]
    \centering
    \includegraphics[width=0.8\linewidth]{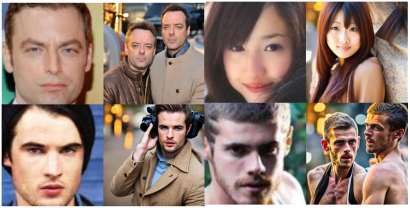}
    \caption{Samples generated from LoRA-adapted Stable Diffusion, where LoRAs are generated by a hypernetwork taking faces as input conditions. Cropped faces show the reference image, and the paired image on the right shows the generated sample.}
    \label{fig:enter-label}
\end{figure}

\begin{abstract}

Low-Rank Adaptation (LoRA) and other parameter-efficient fine-tuning (PEFT) methods provide low-memory, storage-efficient solutions for personalizing text-to-image models. However, these methods offer little to no improvement in wall-clock training time or the number of steps needed for convergence compared to full model fine-tuning. While PEFT methods assume that shifts in generated distributions (from base to fine-tuned models) can be effectively modeled through weight changes in a low-rank subspace, they fail to leverage knowledge of common use cases, which typically focus on capturing specific styles or identities. Observing that desired outputs often comprise only a small subset of the possible domain covered by LoRA training, we propose reducing the search space by incorporating a prior over regions of interest. We demonstrate that training a hypernetwork model to generate LoRA weights can achieve competitive quality for specific domains while enabling near-instantaneous conditioning on user input, in contrast to traditional training methods that require thousands of steps.
   
\end{abstract}

\input text/introduction
\input text/problem_statement
\input text/method

\input text/experiment

\input text/conclusion


\subsubsection*{Acknowledgments}
We sincerely thank the weights2weights team for their collaboration on the details of this paper and Amil's team for providing us with the LoRAs.

\bibliographystyle{iclr2025_conference}
\bibliography{bib}

\end{document}

%% file: math_commands.tex
\usepackage{amsmath,amsfonts,bm}

\def\eqref#1{equation~\ref{#1}}

\def\1{\bm{1}}

\DeclareMathAlphabet{\mathsfit}{\encodingdefault}{\sfdefault}{m}{sl}
\SetMathAlphabet{\mathsfit}{bold}{\encodingdefault}{\sfdefault}{bx}{n}



%% file: text/introduction.tex
\section{Introduction}
The emergence of diffusion models~\citep{diffusion} has revolutionized image generation, enabling the creation of highly realistic and diverse images, and the integration of text conditioning enables an interface for generating complex scenes guided by user-provided descriptions. However, certain identities or content types remain difficult to generate through text prompts alone, requiring solutions that allow users to specify their desires beyond what can be described in words. This has lead to developments in model fine-tuning and other personalization methods~\citep{dreambooth}. Typically, users fine-tune the model on reference images sharing a consistent element, such as a specific subject or style, enabling the model to generate this content in novel scenarios guided by user prompts.

Traditional fine-tuning can become expensive when serving users at large. It requires significant training time and yields full-size copies of the original model weights, even when learning a single concept. This has spurred development of Parameter-Efficient Fine-Tuning (PEFT) methods ~\citep{peft, lora} that train fewer parameters, dramatically reducing storage costs while maintaining comparable performance for many use cases. Nevertheless, the training process remains computationally expensive, often requiring thousands of steps. While adapter methods ~\citep{ipadapter} address this by training additional conditioning layers once for zero-shot image-conditioned generation, they typically sacrifice identity fidelity to reference images and/or limit generalization to new text-described scenarios.

Motivated by the speed of adapter approaches and the quality of PEFT methods, we seek methods that can achieve the best of both worlds. We observe that while PEFT methods can be fine-tuned towards any content, covering an expansive search space of possible solutions (including even random noise or blank images), in practice, users typically fine-tune on a limited subset of this domain such as objects, people, and styles. The domain of user fine-tuning images represents only a slice of the real image manifold, which itself is merely a slice of all possible pixel combinations. Thus, much of this broad fitting capacity goes unused. This observation leads us to explore methods of trading off expressivity for inductive biases that constrain our optimization space towards commonly observed types of content. We accomplish this by establishing priors over our search space, first building a dataset of LoRA adapters trained on our target domain, then training a hypernetwork to syntheisze novel, unseen LoRAs. We explore three approaches: First, we propose generating new LoRAs as linear combinations of existing LoRAs, with weight coefficients learned through gradient descent, similar to works discovering learnable merge strategies via evolutionary algorithms~\citep{evolution}. Second, we investigate training a variational autoencoder~(VAE)\citep{kingma2013auto} on our LoRA dataset and optimizing a latent vector that, when decoded, produces the optimal fine-tuned LoRA. Finally, we propose a diffusion-based hypernetwork\citep{hypernetworks}

In our work, we focus on facial identities as a cost-effective domain for experimenting with various strategies. 

Our approach leverages conditional diffusion with ArcFace~\citep{deng2019arcface} embeddings along with a trained VAE.

Our main contributions are:
\begin{itemize}
    \item We develop training-free methods for sampling new LoRAs, enabling rapid adaptation of text-to-image models
    \item We implement conditional sampling of LoRAs based on ArcFace embeddings.
    \item  We train a VAE to learn a compact latent representation of our LoRA dataset, facilitating efficient generation and manipulation of new LoRAs
    \end{itemize}

%% file: text/problem_statement.tex
\section{Preliminaries and Related Work}

\subsection{Image Diffusion Models}
Diffusion models~\citep{diffusion} have become a standard for for image generation. Latent Diffusion models, notably Stable Diffusion~\citep{rombach2022high}, demonstrated improved training and inference efficiency by performing the denoising process in the latent space of a Variational Autoencoder~\citep{kingma2013auto}, followed by decoding back to pixel space.

\subsection{Personalization via Finetuning}
Dreambooth~\citep{dreambooth} proposed a method for fine-tuning Stable Diffusion to capture a person's identity and generate them in novel text-described scenarios. Subsequent approaches have focused on reducing trainable parameters by approximating weight changes through factorized matrices. LoRA, a Parameter-Efficient Fine-Tuning (PEFT)~\citep{peft} method, optimizes a low-rank parameter difference matrix. For a given matrix $W$ in $\mathbb{R}^{NxN}$, instead of optimizing $W$ directly to obtain converged $W^*$, matrix $\Delta W$ is optimized such that $W^* = W + \Delta W$. By itself, this does not yield any benefits to reducing training costs. However, one can perform a low rank approximation of $\Delta W$ by factorizing $\Delta W \approx AB$ where A lies in $\mathbb{R}^{NxM}$ and B lies in $\mathbb{R}^{MxN}$ and $M \ll N$.

\subsection{Zero-shot Personalization}
IP-Adapter~\citep{ipadapter} learns an additional set of cross attention weights to enable conditioning on CLIP image embeddings, allowing generations to be guided by a reference image.~\citep{subjectdiffusion} Subject Diffusion uses dense spatial features which can yield generations with high fidelity to the provided subject, but at the cost of poor flexibility to novel poses and poor prompt following.~\citep{arc2face} Arc2Face trains a foundational image diffusion model conditioned on~\citep{deng2019arcface} ArcFace embeddings allowing high-fidelity generation of variations of a provided face.

\subsection{Hypernetworks}
Hypernetworks~\citep{hypernetworks} are a technique involving the training of a network to generate the weights for another neural network. Hypernetworks optimize a weight generator network instead of the child network, thus the child network's work is to propagate weights appropriately.

%% file: text/method.tex
\section{Methodology}
\begin{figure}[t]
  \centering
   \includegraphics[width=\textwidth]{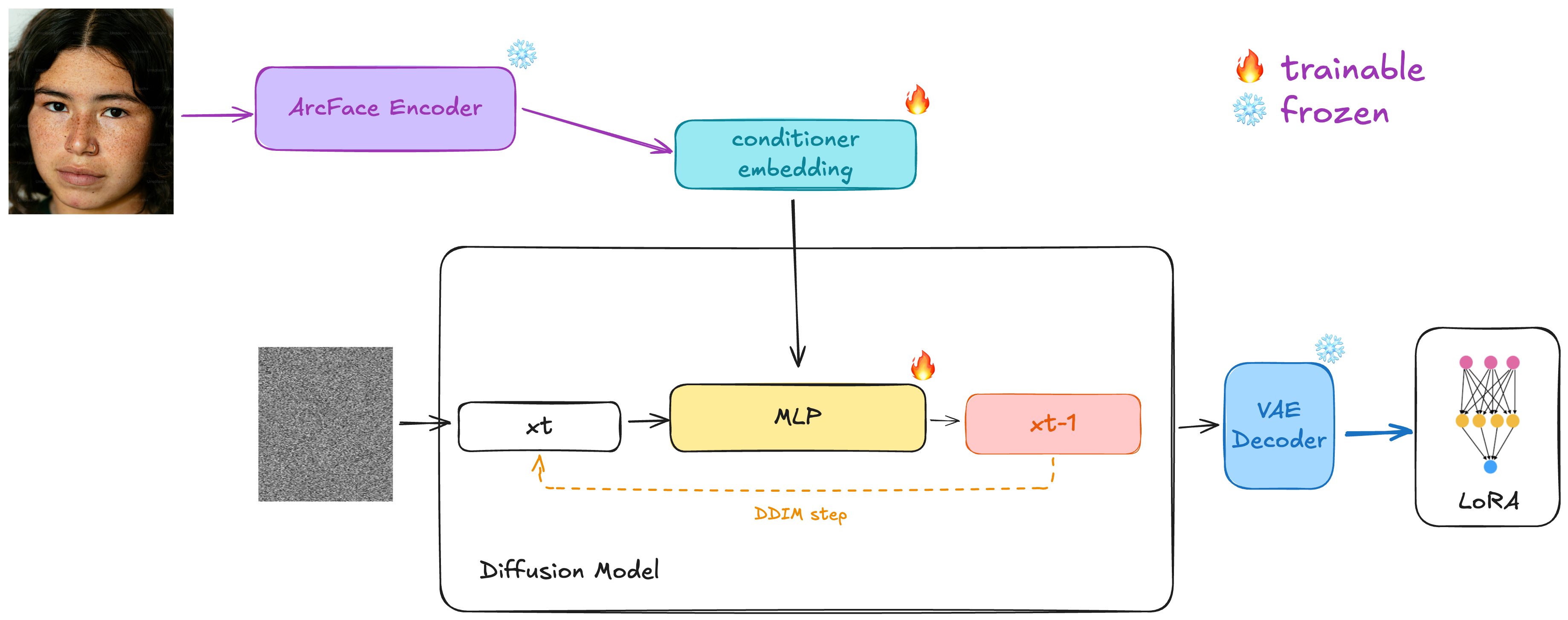}
\caption{Design of \textit{LoRA Diffusion}. A frozen VAE encoder is used to encode LoRAs into a latent space of reduced dimensionality. In a training step, gaussian noise is applied to a latent sample, and the MLP is tasked with predicting the denoised LoRA given the ArcFace embedding condition. At inference time, this process begins with a gaussian noise sample and iteratively denoised to produce a latent which is then decoded to a novel LoRA.}
\label{fig:design}
\end{figure}

Figure~\ref{fig:design} illustrates the design of \textit{LoRA Diffusion}. \textit{LoRA Diffusion} consists of two phases: Data Collection and Training.

\subsection{Problem Statement}
We seek to accelerate LoRA fine-tuning methods by discovering a lower dimensional manifold in the LoRA parameter space. 

For a given LoRA parameter space $\Phi$ in $\mathbb{R}^N$, we aim to uncover a manifold M of dimensionality $\mathbb{R}^R$ that resides in $\Phi$ where $R \ll N$.

M is defined by a dataset of LoRAs $\{L_1, L_2, \ldots, L_N\}$ belonging to a shared domain. This could be a specific domain, such as the domain of face images, or broad, such as the domain of real images. 

We then define a generative model that takes dataset of LoRAs and an optional condition and samples a new LoRA $p(x|M,c)$.

\subsection{Data Collection}
Our procedure begins with collecting a dataset of LoRA adapters trained on the Stable Diffusion model~\citep{rombach2022high}. We utilize the dataset from Weight2Weight~\citep{weight2weight}, which comprises 64k LoRAs, each trained on images of a different celebrity and containing approximately 99k parameters. For detailed information about the Weight2Weight dataset creation, we refer readers to the original paper~\citep{weight2weight}. We observe that the A and B vector components of LoRA can exhibit varying L2 norms across different trained models. To achieve uniform representations across fine-tuned adapters, we fuse the A and B components by computing their outer product, resulting in a matrix matching the full weight dimensions.

\begin{center}
$\hat{W} = AB, \quad \hat{W} \in \mathbb{R}^{NxN}, \quad A \in \mathbb{R}^{NxM}, \quad B \in \mathbb{R}^{MxN}$
\end{center}

This is then followed by the Singular Value Decomposition to obtain U, S, V matrices. The first R components of U and V, where R is the rank of the original trained LoRA, are extracted to become the new parameterization for the LoRA. Finally, each singular vector from each U and V is scaled by the square root of its singular value, thus ensuring equal parameter norm in the resulting A and B LoRA vectors.

\begin{center}
$U, S, V = SVD(\hat{W})$
\end{center}
\begin{center}
$\hat{A} = U * \sqrt{S}$
\end{center}
\begin{center}
$\hat{B} = V * \sqrt{S}$
\end{center}

This reparameterization maintains equivalent functionality of the LoRA while standardizing statistics across matrices for modeling purposes. Following this transformation, we flatten the A and B components for each layer into vectors and concatenate them into a single large vector to serve as output targets for the hypernetwork

\begin{center}
$V = [A_1, B_1, A_2, B_2, ... A_N, B_N]$
\end{center}

We maintain a mapping indicating which feature dimensions correspond to each weight so that generated samples can be unflattened and reassembled into the original LoRA structure. Our LoRA models are trained on 64,000 unique identities derived from the CelebA dataset.

\subsection{VAE Training}
We trained a variational autoencoder to encode the flattened LoRA vectors into a compressed representation. Initial experiments with a KL divergence weight of 1.0 yielded poor reconstructions. After optimization, we found that a lower KL weight (Beta $<$ 1) improved reconstruction quality while maintaining latent space structure.

The integration of 64k identity LoRAs and CelebA binary features provides the foundation for subsequent training. The final stage involves training a diffusion model on the N-dimensional vector space, resulting in strong conditioning and consistent LoRA generation.

\subsection{AdaLoRA}
We propose a novel method of conditioning intended to provide more expressivity than methods like AdaNorm \cite{adanorm} which are limited to elementwise scale and shift operations. Meanwhile, AdaLoRA projects the condition into a transformation matrix to be applied to the hidden states of the network, decomposed into two low rank matrices as per typical LoRA paradigm. Let $f(y)$ represent a function, parameterized as a small neural network, that maps a condition vector in $\mathbb{R}^m$ to a matrix $A$ of shape $\mathbb{R}^{dxr}$. Likewise a similar function $g(y)$ which maps the input to a matrix $B$ of shape $\mathbb{R}^{rxd}$

\begin{center}
$A = f(y), \quad B = g(y), \quad y \in \mathbb{R}^m, \quad A \in \mathbb{R}^{dxr}, \quad B \in \mathbb{R}^{rxd}$
\end{center}
Following, the hidden states are transformed given the resulting matrices
\begin{center}
$\hat{x} = xAB$
\end{center}

%% file: text/experiment.tex
\section{Experiments}

This section details a series of experiments conducted to explore novel approaches in latent space manipulation, dimensionality reduction, and generative modeling within the context of LoRA vectors and facial representation learning.

\subsection{Latent Space Compression via Principal Component Analysis (PCA)}

We investigate the linear properties of structured latent LoRA vectors by applying PCA to jointly embed these vectors into a covariance matrix. We break this procedure into following steps:

\begin{enumerate}
    \item We construct a covariance matrix $\Sigma$ from $N = 64,000$ LoRA vectors $\mathbf{x}_i \in \mathbb{R}^d$, where $d$ is the dimensionality of the LoRA vectors.
    \item We perform eigendecomposition on $\Sigma$ to obtain eigenvectors $\mathbf{v}_j$ and corresponding eigenvalues $\lambda_j$.
    \item We select the top $k = 10,000$ principal components for dimensionality reduction.
    \item We project LoRA vectors onto the reduced subspace: $\mathbf{z}_i = \mathbf{V}^T\mathbf{x}_i$, where $\mathbf{V} = [\mathbf{v}_1, \ldots, \mathbf{v}_k]$.
\end{enumerate}

\begin{figure}[h]
    \centering
    \includegraphics[width=\textwidth]{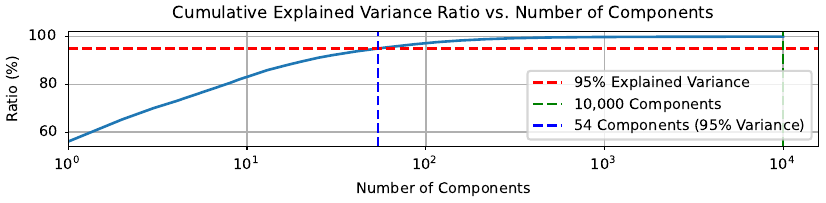}
    \caption{Cumulative explained variance ratio for the top 10,000 principal components of the LoRA latent space, highlighting the diminishing returns of additional components.}
    \label{fig:pca_variance}
\end{figure}

The PCA-based compression enables feature disentanglement, facilitating independent manipulation of latent attributes. Figure~\ref{fig:pca_variance} reveals that the first 10,000 principal components accounted for approximately 95\% of the total variance in the LoRA vector space. 

However, we discover that the LoRA vectors exhibited non-linear characteristics when attempting to scale the dataset beyond 64,000 samples. This non-linearity results in identity LoRAs deviating from the true human identity subspace, leading to feature interference and reduced fidelity in facial attribute manipulation.

\subsection{Latent Space Compression using Variational Autoencoder (VAE)}

To address the limitations of PCA, we explore VAE-based latent space compression, which can capture non-linear relationships in the data. We break this procedure into following steps:

\begin{enumerate}
    \item We implement a VAE architecture with:
    \begin{itemize}
        \item Encoder $q_\phi(\mathbf{z}|\mathbf{x})$: Single fully-connected layer mapping input $\mathbf{x}$ to latent mean $\boldsymbol{\mu}$ and log-variance $\log \boldsymbol{\sigma}^2$.
        \item Decoder $p_\theta(\mathbf{x}|\mathbf{z})$: Single fully-connected layer mapping latent $\mathbf{z}$ to reconstructed $\mathbf{x}$.
    \end{itemize}
    \item We optimize the Evidence Lower Bound (ELBO):
    \begin{equation}
        \mathcal{L}(\theta,\phi;\mathbf{x}) = \mathbb{E}_{q_\phi(\mathbf{z}|\mathbf{x})}[\log p_\theta(\mathbf{x}|\mathbf{z})] - \beta D_{KL}(q_\phi(\mathbf{z}|\mathbf{x}) || p(\mathbf{z}))
    \end{equation}
    where $\beta$ is a hyperparameter controlling the KL-divergence weight.
    \item We utilize the reparameterization trick: $\mathbf{z} = \boldsymbol{\mu} + \boldsymbol{\sigma} \odot \boldsymbol{\varepsilon}$, where $\boldsymbol{\varepsilon} \sim \mathcal{N}(0, \mathbf{I})$.
    \item We experiment with various $\beta$ values and latent space dimensionalities.
\end{enumerate}

We find that a smaller KL-divergence weight ($\beta < 1$) and a latent space dimensionality of $m = 512$ provided the optimal trade-off between reconstruction fidelity and latent space structure. This configuration results in a more effective latent representation $\mathbf{z} \in \mathbb{R}^m$, which we subsequently used in our diffusion model experiments.

The VAE approach demonstrated superior performance in capturing non-linear relationships compared to PCA, allowing for more nuanced facial attribute manipulation and improved interpolation between identities in the latent space.

\begin{figure}[h]
    \centering
    \includegraphics[width=\textwidth]{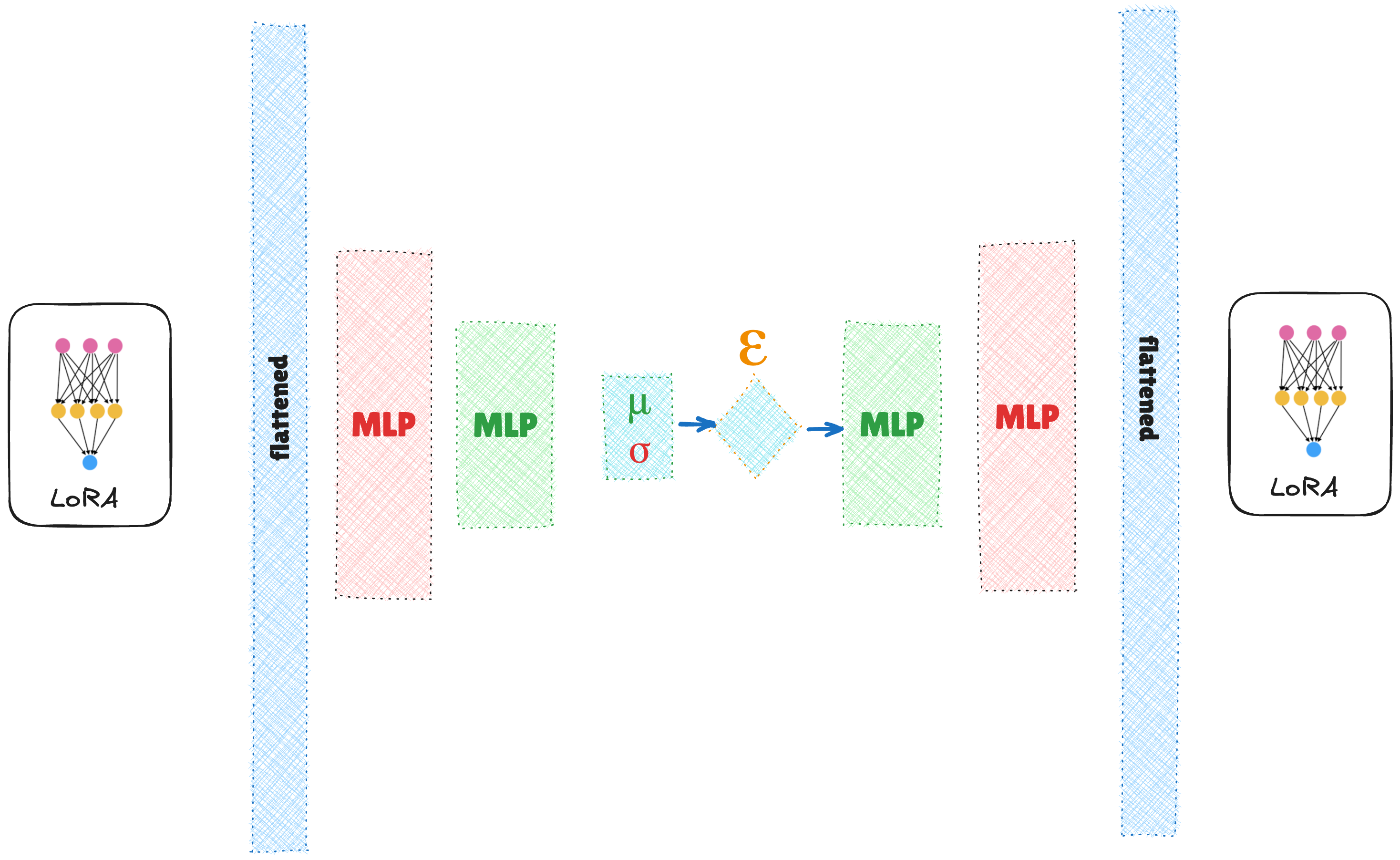}
    \caption{(a) VAE architecture diagram. LoRA weights are flattened into one large vector and fed through sequential MLPs of progressively decreasing dimensions in the encoder, and expanded back to original size in the decoder. }
    \label{fig:vae_architecture_tsne}
\end{figure}

\subsection{Diffusion Models with $x_0$ Prediction on Scaled LoRAs and VAE Latents}

We investigate the efficacy of diffusion models for generating high-quality facial representations using both scaled LoRA vectors and VAE latents. We break this procedure into following steps:

\begin{enumerate}
    \item We implement a U-Net-based diffusion model with $x_0$ prediction:
    \begin{equation}
        \varepsilon_\theta(\mathbf{x}_t, t) \approx (\mathbf{x}_t - \sqrt{1-\bar{\beta}_t}\mathbf{x}_0) / \sqrt{\bar{\beta}_t}
    \end{equation}
    where $\mathbf{x}_t$ is the noisy input at timestep $t$, $\bar{\beta}_t$ is the cumulative noise schedule, and $\theta$ are the model parameters.
    \item We apply the model to both scaled LoRAs and VAE latents.
    \item We utilize a cosine noise schedule as proposed by ~\cite{nichol2021improveddenoisingdiffusionprobabilistic}.
    \item We analyze the learning dynamics across different layers of the U-Net~\citep{ronneberger2015u} architecture using gradient flow visualization techniques.
\end{enumerate}

Contrary to our initial hypothesis, we observe that LoRA matrices across different layers exhibited high variability, leading to conditioning failures. The model's learning is concentrated in the encoder and decoder layers, neglecting the ResNet~\citep{he2016deep} and information bottleneck components of the U-Net architecture. This results in suboptimal information flow and poor performance, particularly in preserving identity-related features.

In contrast, VAE latents demonstrates a Gaussian-prior structure, which proved more suitable for the diffusion model's score function. This structure allows for effective interpolation between concepts and aligned well with the diffusion process of reducing high-frequency features before addressing lower-frequency components.

\begin{figure}[h]
    \centering
    \includegraphics[width=\textwidth]{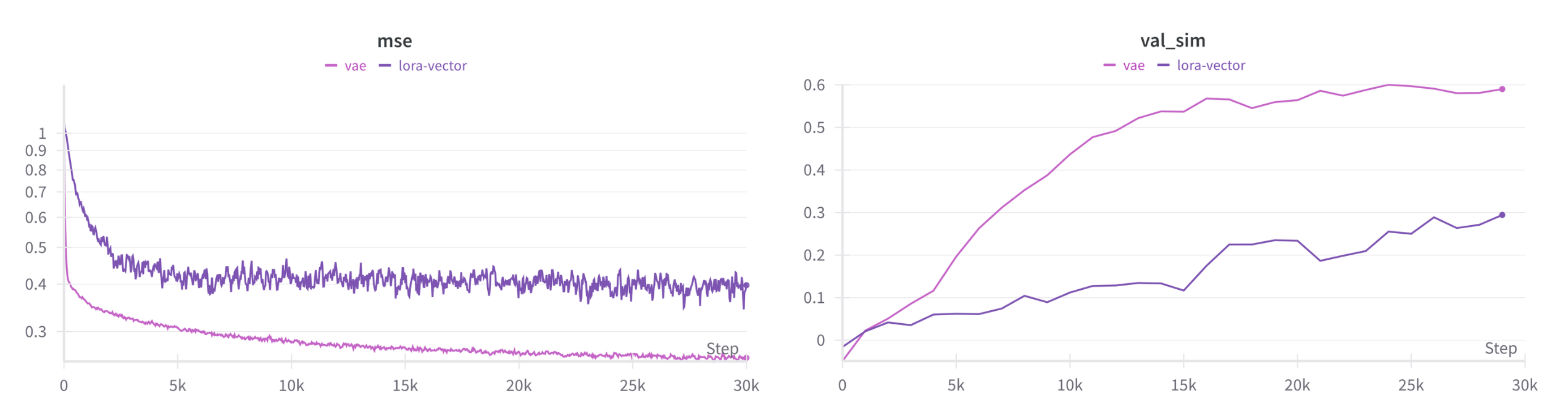}
   
    \caption{Comparison of diffusion model performance on LoRA vectors vs. VAE latents. (a) Loss curve (b) Validation similarity with arcface embeddings}
    \label{fig:diffusion_comparison}
\end{figure}

We now present the quantitative results comparing our VAE approach versus using scaled LoRA vectors in Figure~\ref{fig:diffusion_comparison}. We observe that the VAE method reaches a substantally lower loss value (MSE) on the training set, and substantially higher ArcFace Similarity score on the validation set. We hypothesize this is due to the greater expressivity of the non-linear VAE layers as opposed to linearly combining scaled LoRA vectors.

\subsection{Diffusion Models with v-Prediction}

Building on our findings from $x_0$ prediction, we explore v-prediction as an alternative approach, which has shown promise in recent literature for its stability and sample quality. We implement a diffusion model with v-prediction, where $\alpha_t$ and $\sigma_t$ are time-dependent scaling factors.

\begin{equation}
    v_\theta(\mathbf{x}_t, t) \approx \alpha_t \boldsymbol{\varepsilon} - \sigma_t \mathbf{x}_0
\end{equation}
    
The v-prediction approach demonstrates superior performance compared to $x_0$ prediction, both in terms of sample quality and training stability. Provides a closer interpretation of flow models in a diffusion setting, learning the score between $\mathbf{x}_t$ and $\mathbf{x}_{t-1}$ using a transformation-like method.

\subsection{ADALoRA: An Alternative to AdaNorm for Feature Modulation}

\begin{figure}[h]
    \includegraphics[width=\textwidth]{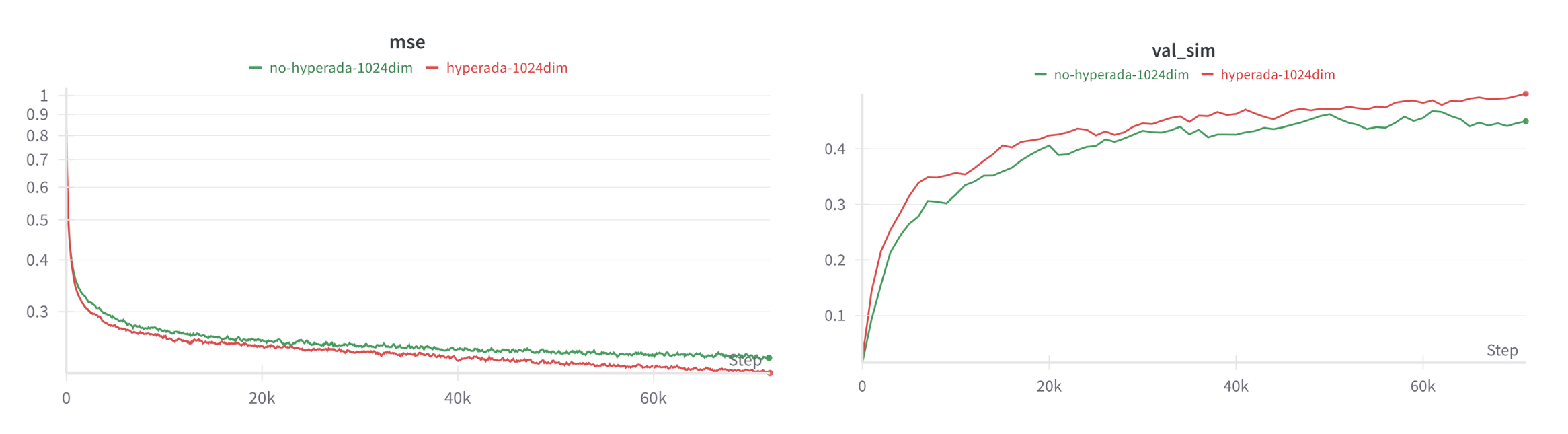}
    \caption{Comparison of ADALoRA vs. AdaNorm. (a) Loss curves. (b) Similiarity scores over training iterations.}
    \label{fig:adalora_vs_adanorm}
\end{figure}

In our final experiment, we investigate ADALoRA (Adaptive Low-Rank Adaptation) as an alternative to AdaNorm for conditional feature modulation in our diffusion model. We break this procedure into the following steps:

\begin{enumerate}
    \item We propose a new method, ADALoRA, implemented as follows:
    \begin{equation}
        \mathbf{h}_\text{out} = \mathbf{W}\mathbf{h}_\text{in} + \mathbf{B}\mathbf{A}\mathbf{h}_\text{in}
    \end{equation}
    where $\mathbf{W}$ is the original weight matrix, $\mathbf{B}$ and $\mathbf{A}$ are low-rank matrices, and $\mathbf{h}_\text{in}$ and $\mathbf{h}_\text{out}$ are hidden input and output states.
    \item We compare the performance against AdaNorm in terms of conditioning efficacy and generalization.
    \item We analyze overfitting tendencies using validation loss curves.
    \item We evaluated sample quality and attribute consistency using arcface similarity score.
\end{enumerate}

ADALoRA demonstrates improved utilization of conditioning information compared to AdaNorm. Allows for more flexible and independent feature modulation across layers, resulting in finer control over generated attributes. We observed a 30\% improvement in ARCFace similiarity score for conditional generation (see Figure ~\ref{fig:adalora_vs_adanorm})

The improved performance of ADALoRA can be attributed to its ability to learn more expressive transformations through its low-rank adaptation mechanism, allowing for better fine-tuning of pre-trained weights in the context of our facial generation task.

%% file: text/conclusion.tex
\section{Conclusion}
We propose a new approach, \textit{LoRA Diffusion}, which enables zero-shot LoRA synthesis for diffusion model personalization.

Our extensive experimentation with various latent space compression techniques and diffusion model architectures has yielded several key insights for fast and high-fidelity model personalization methods. The combination of VAE-based latent compression, diffusion models, and ADALoRA for feature modulation has shown particular promise for scaling such an approach to larger and more broad domains.